\documentclass{tx}
\usepackage[utf8]{inputenc}
\usepackage{natbib}
\usepackage{overpic}
\usepackage{xcolor}
\usepackage{amssymb}
\usepackage{booktabs,makecell,multirow}
\usepackage{pgfplots}
\pgfplotsset{compat=1.18}
\usetikzlibrary{arrows.meta,pgfplots.colormaps,positioning,decorations.pathreplacing}
\usepgfplotslibrary{groupplots}

\usepackage[most]{tcolorbox}
\usepackage{fontawesome5}
\newcommand{\frozen}{{\color{cyan}\faSnowflake}}
\newcommand{\trainable}{{\color{orange}\faFire}}

\usepackage{colortbl}
\usepackage{arydshln}  
\definecolor{rowgray}{gray}{0.93}

\definecolor{ourscolor}{RGB}{30,78,140}   
\definecolor{lewmcolor}{RGB}{176,125,44}    

\usepackage{hyperref}
\hypersetup{
  colorlinks=true,
  linkcolor=blue!70!black,
  citecolor=blue!70!black,
  urlcolor=blue!70!black,
}
\usepackage{cleveref}
\crefname{figure}{Figure}{Figures}
\crefname{section}{Section}{Sections}
\crefname{table}{Table}{Tables}
\crefname{equation}{Equation}{Equations}
\crefname{appendix}{Appendix}{Appendices}
\crefname{theorem}{Theorem}{Theorems}
\crefname{lemma}{Lemma}{Lemmas}
\crefname{definition}{Definition}{Definitions}
\crefname{corollary}{Corollary}{Corollaries}
\crefname{proposition}{Proposition}{Propositions}
\crefname{example}{Example}{Examples}

\RequirePackage{orcidlink}

\newcommand{\ds}{d_s}
\newcommand{\z}{\boldsymbol{z}}
\newcommand{\hatz}{\hat{\boldsymbol{z}}}
\newcommand{\proj}{\mathbf{P}}
\newcommand{\Z}{\mathbf{Z}}

\usepackage[most]{tcolorbox}

\newcommand{\name}{Sub-JEPA}

\title{
    \name: Subspace Gaussian Regularization for Stable End-to-End World
    Models
}
\author{
    Kai Zhao\textsuperscript{1}\orcid{0000-0002-2496-0829},
    Dongliang Nie\textsuperscript{1},
    Yuchen Lin\textsuperscript{1},
    Zhehan Luo\textsuperscript{2}, 
    Yixiao Gu\textsuperscript{1},
    Deng-Ping Fan\textsuperscript{3}\orcid{0000-0002-5245-7518},
    Dan Zeng\textsuperscript{1}
}
\affiliation{
    \textsuperscript{1}Shanghai University \quad \textsuperscript{2}The University of Manchester \quad \textsuperscript{3}Nankai University
}
\email{
    \url{https://github.com/intcomp/sub-jepa}
}

\begin{abstract}
Joint-Embedding Predictive Architectures (JEPAs) provide a simple
framework for learning world models by predicting future latent representations.
However, JEPA training is subject to a bias-variance tradeoff.
Without sufficient structural constraints, excessive representational
variance causes the model to collapse to trivial solutions.
The recent LeWorldModel (LeWM) shows that this issue can be alleviated by
simply constraining latent embeddings with an isotropic Gaussian prior.
However, latent representations inherently lie on low-dimensional manifolds
within a high-dimensional ambient space, and enforcing an isotropic Gaussian
prior directly in this ambient space introduces an overly strong bias.
In this work, we propose \name, which seeks a favorable operating
point on the bias-variance frontier by applying Gaussian constraints in
multiple random subspaces rather than in the original
embedding space.
This design relaxes the global constraint while preserving its
anti-collapse effect, leading to a better balance between training
stability and representation flexibility.
Extensive experiments across four
continuous-control environments demonstrate that \name{} consistently
outperforms LeWM with very clear margins.
Our method is simple yet effective, and can serve as a strong baseline for future JEPA-based world model research.
\ifdefined\ieeemode
The code is available at \url{https://github.com/intcomp/Sub-JEPA}.
\fi
\end{abstract}
\begin{document}
\maketitle


\section{Introduction}\label{sec:intro}
World models (WM)~\cite{ha2018world,hafner2025mastering},
predictive representations of how
environments evolve under
actions,
have become critical building
blocks of modern artificial intelligence.
WM provides a principled route of learning compact
representations of the observations and predicting future states under candidate actions.
Such predictive models are attractive because they support planning in latent
space while avoiding the cost of modeling every pixel in the 
observations~\cite{hafner2025mastering,iris2023}.

Among recent approaches, 
the Joint-Embedding Predictive Architecture (JEPA)
has emerged as a particularly appealing formulation for latent world
modeling~\cite{lecun2022path}.
JEPA learns an encoder that maps observations to latent representations and 
a dynamics model that predicts future latent states conditioned on the current 
representation and action~\cite{sobal2022jointembeddingpredictivearchitectures}.
This design shifts modeling capacity toward task-relevant abstractions 
and avoids the burden of full observation reconstruction~\cite{assran2023self}.

Despite this conceptual simplicity,
stable end-to-end training of JEPA-based models is
subject to a bias-variance tradeoff~\cite{jingunderstanding,sobal2022jointembeddingpredictivearchitectures}:
insufficient constraints lead to representation collapse,
while excessive constraints suppress the richness of learned representations.
Early methods attempted to mitigate collapse 
by introducing various 
regularization terms~\cite{bardes2022vicreg,sobal2025stresstesting}
or heuristic training recipes~\cite{assran2023self,bardes2024revisiting,zhou2025dino-wm}, 
but these approaches are often complex,
not end-to-end, and lack theoretical guidance.
%
%
The LeWorldModel (LeWM)~\cite{maes_lelidec2026lewm} pioneers stable end-to-end training
of JEPA-based world models by introducing a simple bias
that regularizes latent embeddings toward an 
isotropic Gaussian distribution.
%
%

However, the latent representations of natural control tasks 
typically lie on low-dimensional manifolds embedded within the 
high-dimensional ambient space~\cite{bengio2013representation,tenenbaum2000global},
and enforcing an isotropic Gaussian prior directly in this ambient space
introduces an overly strong bias that mismatches the intrinsic geometry
of the underlying dynamics~\cite{maes_lelidec2026lewm}.

In this work, we propose \name, which seeks a more favorable operating
point on the bias-variance frontier by moving Gaussian regularization
from the ambient space into low-dimensional subspaces.
%
%
This simple design relaxes the global isotropic prior 
while preserving its anti-collapse effect,
yielding a bias better matched to the intrinsic dimensionality of the underlying dynamics.
%
%
\name{} retains the simplicity of the LeWM training recipe
while introducing a more flexible structural prior over the latent space.
%
%
Our main contributions are as follows:
\begin{itemize}
    \item We identify that the global isotropic Gaussian prior in LeWM 
    introduces an excessive bias for tasks with low intrinsic dimensionality.
    \item We propose \name, which relocates Gaussian regularization
    from the ambient embedding space to multiple low-dimensional
    row-orthonormal projected views, effectively relaxing the global
    prior while preserving the anti-collapse effect.
    \item Extensive experiments across four continuous-control environments 
    demonstrate that \name{} consistently outperforms LeWM by clear margins,
    with gains correlated with reductions in effective rank,
    confirming that subspace regularization better respects 
    the intrinsic dimensionality of the underlying dynamics.
\end{itemize}

\ifdefined\ieeemode
  \IEEEpubidadjcol
\fi

\section{Related Work}\label{sec:related}
\textbf{Latent World Models.}
World models learn a compressed model of environment dynamics from observations, enabling agents to plan without repeated real-world interaction~\cite{ha2018world}.
Early approaches predict in pixel space, which is computationally expensive and burdens the model with irrelevant visual details~\cite{lecun2022path}.
Autoregressive generative models such as IRIS~\cite{iris2023} and DreamerV3~\cite{hafner2025mastering} couple the world model with an image decoder and achieve strong results in reward-driven settings, but reconstruction-based objectives can produce embeddings that are uninformative for control.
Latent-space world models instead predict future embeddings directly, avoiding reconstruction cost while concentrating capacity on task-relevant structure~\cite{lecun2022path}.
Test-time planning via Model Predictive Control over such latent dynamics models has demonstrated strong performance across continuous-control and navigation tasks~\cite{Hansen2022tdmpc, zhou2025dino-wm, sobal2025stresstesting}, but requires a well-structured, non-degenerate latent space.

\textbf{JEPA and End-to-End Training}.
Joint-Embedding Predictive Architectures~\cite{lecun2022path} train an encoder and a predictor jointly to anticipate the embedding of a future or masked view, without reconstructing raw pixels.
This recipe has been instantiated in image representation learning with I-JEPA~\cite{assran2023self}, in video modeling with V-JEPA~\cite{bardes2024revisiting}, and in action-conditioned latent dynamics~\cite{sobal2022jointembeddingpredictivearchitectures, zhou2025dino-wm}.
The central challenge is representation collapse~\cite{jingunderstanding}: without explicit structural constraints the encoder can map all inputs to nearly identical embeddings, trivially minimizing the prediction loss while destroying useful structure.
Existing latent world model solutions avoid collapse either by using pretrained frozen encoders such as DINOv2~\cite{zhou2025dino-wm}, which limits representation flexibility, or by adding multi-term collapse-prevention losses, as in PLDM~\cite{sobal2025stresstesting}, which applies VICReg~\cite{bardes2022vicreg} and requires tuning several sensitive hyperparameters.
LeWorldModel (LeWM)~\cite{maes_lelidec2026lewm} achieves stable end-to-end training from raw pixels using only two loss terms: a next-embedding prediction objective and Gaussian regularizer.

\textbf{Anti-Collapse Regularization}. A broad class of self-supervised methods prevents collapse by imposing structure on the embedding distribution.
Contrastive methods such as SimCLR~\cite{chen2020simple} and MoCo~\cite{he2020momentum} push apart embeddings of different samples but require large batches or memory banks.
Non-contrastive approaches such as BYOL~\cite{grill2020bootstrap} rely on teacher-student asymmetry with stop-gradient, while Barlow Twins~\cite{zbontar2021barlow} and VICReg~\cite{bardes2022vicreg} decorrelate embedding dimensions; Whitening-MSE~\cite{ermolov2021whitening} further enforces a uniform distribution on the unit sphere.
Gaussian regularization, introduced in LeJEPA~\cite{balestriero2025lejepa}, takes a principled approach: random one-dimensional projections of the embeddings are tested for Gaussianity, and by the Cramér-Wold theorem~\cite{cramer} matching all projected marginals to a Gaussian implicitly enforces an isotropic Gaussian joint distribution.
LeJEPA proves this prior to be the unique optimal embedding distribution for minimizing downstream prediction risk, placing the collapse-prevention objective on a rigorous theoretical footing.

\textbf{Subspace Methods and Random Projections.}
Random projections underpin scalable dimensionality reduction via the Johnson-Lindenstrauss lemma~\cite{halko2011finding} and enable tractable distribution matching through sliced Wasserstein distances~\cite{bonneel2015sliced}, which reduce high-dimensional optimal transport to a sequence of one-dimensional comparisons.
Gaussian regularization itself belongs to this family, sketching the embedding distribution with random directions to bypass the curse of dimensionality.
In representation learning, orthogonal projections have been used to decorrelate and spread features~\cite{ermolov2021whitening}.
\name{} extends this principle to Gaussian regularization by applying the regularizer over multiple fixed random row-orthonormal projections of the latent representation.
Instead of enforcing a single global isotropic prior in the ambient space, the regularization is imposed independently on multiple low-dimensional projected views, yielding a more flexible inductive bias that better matches the intrinsic structure of the underlying dynamics.

\section{Method}\label{sec:method}

In this section, we formally introduce our subspace-regularized latent world
model for learning from state-action trajectories without reward annotations.
We first present the problem setting of latent dynamics learning, and then
describe our proposed \name{} framework, which extends LeWorldModel (LeWM)~\cite{maes_lelidec2026lewm} by
replacing full-space Gaussian regularization with Gaussian regularization over
multiple frozen row-orthonormal random projections.

\subsection{Problem Setup}
\label{subsec:setup}

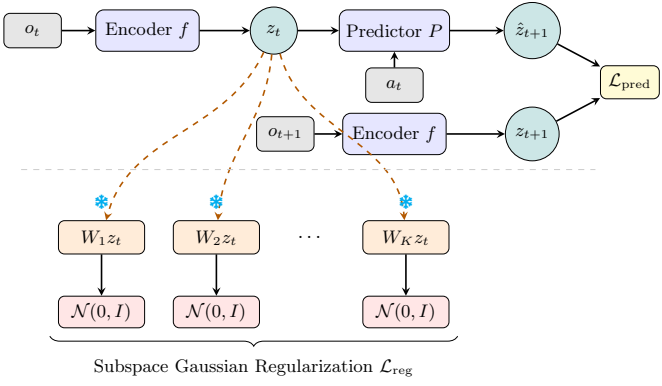
\begin{figure}[t]
\centering
\resizebox{\linewidth}{!}{%
\begin{tikzpicture}[
    box/.style={draw, rounded corners=4pt, minimum width=1.8cm, minimum height=0.72cm,
                align=center, fill=blue!10, font=\small},
    obs/.style={draw, rounded corners=3pt, minimum width=1.0cm, minimum height=0.6cm,
                align=center, fill=gray!20, font=\small},
    lat/.style={draw, circle, minimum size=0.82cm, align=center, fill=teal!20, font=\small},
    sub/.style={draw, rounded corners=3pt, minimum width=1.5cm, minimum height=0.58cm,
                align=center, fill=orange!15, font=\small},
    gauss/.style={draw, rounded corners=3pt, minimum width=1.5cm, minimum height=0.52cm,
                  align=center, fill=red!10, font=\small},
    lossn/.style={draw, rounded corners=3pt, fill=yellow!20, font=\small, inner sep=4pt},
    arr/.style={->, >=stealth, thick},
    darr/.style={->, >=stealth, dashed, color=orange!70!black, thick},
]
\node[obs] (ot)   at (0.0,  2.6) {$o_t$};
\node[box] (enc1) at (2.0,  2.6) {Encoder $f$};
\node[lat] (zt)   at (4.2,  2.6) {$z_t$};
\node[obs] (at)   at (6.3,  1.65) {$a_t$};
\node[box] (pred) at (6.3,  2.6) {Predictor $P$};
\node[lat] (zhat) at (8.7,  2.6) {$\hat{z}_{t+1}$};
\node[obs] (ot1)  at (4.4,  0.8) {$o_{t+1}$};
\node[box] (enc2) at (6.3,  0.8) {Encoder $f$};
\node[lat] (zt1)  at (8.7,  0.8) {$z_{t+1}$};
\node[lossn] (Lpred) at (10.4, 1.7) {$\mathcal{L}_{\mathrm{pred}}$};
\node[sub]   (s1)    at (1.2,  -1.0) {$W_1 z_t$};
\node[sub]   (s2)    at (3.2,  -1.0) {$W_2 z_t$};
\node        (sdots) at (4.85, -1.0) {$\cdots$};
\node[sub]   (sk)    at (6.5,  -1.0) {$W_K z_t$};
\node[gauss] (g1)   at (1.2,  -2.3) {$\mathcal{N}(0,I)$};
\node[gauss] (g2)   at (3.2,  -2.3) {$\mathcal{N}(0,I)$};
\node[gauss] (gk)   at (6.5,  -2.3) {$\mathcal{N}(0,I)$};
\draw[arr] (ot)   -- (enc1);
\draw[arr] (enc1) -- (zt);
\draw[arr] (zt)   -- (pred);
\draw[arr] (at)   -- (pred);
\draw[arr] (pred) -- (zhat);
\draw[arr] (ot1)  -- (enc2);
\draw[arr] (enc2) -- (zt1);
\draw[arr] (zhat) -- (Lpred);
\draw[arr] (zt1)  -- (Lpred);
\draw[darr] (zt) to[out=245,in=75] (s1);
\draw[darr] (zt) to[out=265,in=85] (s2);
\draw[darr] (zt) to[out=285,in=95] (sk);
\draw[arr] (s1) -- (g1);
\draw[arr] (s2) -- (g2);
\draw[arr] (sk) -- (gk);
\node[above=2pt of s1, font=\scriptsize] {\frozen};
\node[above=2pt of s2, font=\scriptsize] {\frozen};
\node[above=2pt of sk, font=\scriptsize] {\frozen};
\draw[decorate, decoration={brace, amplitude=5pt, mirror}]
    (0.3, -2.72) -- (7.4, -2.72)
    node[midway, below=8pt, font=\small]
    {Subspace Gaussian Regularization $\mathcal{L}_{\mathrm{reg}}$};
\draw[gray!40, dashed, thin] (-0.2, 0.18) -- (9.8, 0.18);
\end{tikzpicture}%
}
\vspace{-2em}
\caption{Overview of \name{}. Observations $o_t$ and $o_{t+1}$ are encoded by a shared encoder $f$ into latents $z_t$ and $z_{t+1}$. The predictor $P$ maps $(z_t, a_t)$ to $\hatz_{t+1}$, trained with prediction loss $\mathcal{L}_{\mathrm{pred}}$. Below the dashed line, $z_t$ is projected onto $K$ frozen (\frozen) row-orthonormal random projections. $\{W_k\}$; a subspace Gaussian regularization loss $\mathcal{L}_{\mathrm{sub}}$ enforces $\mathcal{N}(0,I)$ in each subspace.}
\label{fig:overview}
\end{figure}

As illustrated in~\cref{fig:overview}, a latent world model consists of
an encoder $f$ and a predictor $P$.
Given a raw observation $o_t$,
the encoder maps it to a compact latent
representation $\z_t = f(o_t) \in \mathbb{R}^D$.
The predictor then estimates the next latent
$\hatz_{t+1} = P(\z_t, a_t)$ conditioned on the current latent and
action $a_t$,
and is trained to match the target latent
$\z_{t+1} = f(o_{t+1})$ obtained by encoding the subsequent observation.
This formulation captures environment dynamics entirely in latent space,
without requiring pixel-level reconstruction.

We consider a fully offline and reward-free setting, where the model is
trained only from pre-collected trajectories of observations and actions,
without access to rewards or task labels, following the JEPA line of work~\cite{assran2025vjepa2selfsupervisedvideo,maes_lelidec2026lewm,zhou2025dino-wm}.
The offline dataset consists of trajectories of length $T$:
\begin{equation}
    \mathcal{D}=\{(o_{1:T}, a_{1:T})\},
\end{equation}
where $o_t$ and $a_t$ denote 
the observation (e.g. an RGB image of the environment), 
and action at time step $t$ (e.g.  continuous control signal),
respectively.

\subsection{Orthogonal Subspace Projection}
\label{subsec:osd}

Let $\z \in \mathbb{R}^{D}$ denote the latent representation produced by the encoder.
We consider a collection of $K$ linear projections that map $\z$ to
low-dimensional representations in $\mathbb{R}^{\ds}$.
Specifically, we introduce
\begin{equation}
    \{\proj_k \in \mathbb{R}^{\ds \times D}\}_{k=1}^{K},
\end{equation}
where $\ds < D$ denotes the subspace dimension.
We set $\ds = \lfloor D / K \rceil$ to reduce the number of hyperparameters
and maintain a simple parameterization.
The projected embedding is defined as
\begin{equation}
    \z^{(k)} = \proj_k \z \in \mathbb{R}^{\ds}, \quad k = 1,\dots,K.
\end{equation}

Each projection matrix $\proj_k$ is constructed by first sampling
a random Gaussian matrix, followed by QR decomposition to obtain an orthonormal basis.
The resulting orthogonal matrix is subsequently transposed to form
the projection matrix $\proj_k \in \mathbb{R}^{\ds \times D}$,
yielding a row-orthonormal projection matrix.
The orthogonalization ensures that the resulting projection directions are mutually 
orthogonal, 
preventing redundancy and preserving geometric structure during projection~\cite{johnson1984extensions}.

Freezing the projections prevents the regularizer itself from adapting
to the evolving latent distribution~\cite{rahimi2007random,saxe2014exact}, 
ensuring that the Gaussian constraint is consistently applied across 
stable and independent subspaces throughout training.

\subsection{Multi-Subspace Gaussian Regularization}
\label{subsec:multisig}

Let $\Z \in \mathbb{R}^{N \times B \times D}$ denote the latent
embedding tensor collected over temporal history length $N$ and
batch size $B$.
%
We apply Gaussian regularization independently in each of
the $K$ row-orthonormal random projection subspaces.
Concretely, let $\{\proj_k \in \mathbb{R}^{\ds \times D}\}_{k=1}^{K}$
denote the orthogonal projection matrices introduced in
Section~\ref{subsec:osd}.
The projected tensor for the $k$-th subspace is obtained by applying
$\proj_k$ along the embedding dimension of $\Z$:
\begin{equation}
    \Z^{(k)} = \Z \proj_k^\top \in \mathbb{R}^{N \times B \times \ds},
    \quad k = 1, \dots, K,
\end{equation}
where each slice $\Z^{(k)}_{n,b,:} \in \mathbb{R}^{\ds}$ is the
$\ds$-dimensional subspace representation of the latent embedding
at temporal index $n$ and batch index $b$.

We then apply Gaussian regularization independently in each subspace.
Following the formulation of LeWM, we sample $M$ random unit vectors
$\{u^{(m)}\}_{m=1}^{M} \subset \mathcal{S}^{\ds-1}$, and compute
one-dimensional projections by taking inner products along the last
dimension:
\begin{equation}
    z_{n,b}^{(k,m)}
    =
    \left\langle \Z^{(k)}_{n,b,:}, \, u^{(m)} \right\rangle,
    \quad n=1,\dots,N,\; b=1,\dots,B.
\end{equation}

Here, the superscript $(k,m)$ indicates that the scalar projection is
computed in the $k$-th subspace along the $m$-th random direction, while
the subscripts $(n,b)$ denote the temporal index and batch index,
respectively. Therefore, $z_{n,b}^{(k,m)}$ is a scalar value obtained
from the latent embedding at temporal index $n$ and batch index $b$.

For each $(k,m)$, 
all projections
$\{z_{n,b}^{(k,m)}\}_{n=1,b=1}^{N,B}$ are scalar samples. 
We evaluate the Epps--Pulley~\cite{epps-pulley} normality
statistic on this sample set:
\begin{equation}
    T^{(k,m)}
    =
    T\!\left(\{z_{n,b}^{(k,m)}\}_{n=1,b=1}^{N,B}\right).
\end{equation}

The regularization objective averages over $M$ random directions and $K$ subspaces:
\begin{equation}
    \mathcal{L}_{\text{reg}}
    =
    \frac{1}{KM}
    \sum_{k=1}^{K}
    \sum_{m=1}^{M}
    T^{(k,m)}.
\end{equation}
The overall training objective is:
\begin{equation}
    \mathcal{L}_{\text{total}} = \mathcal{L}_{\text{pred}}(\hatz_{t+1}, \z_{t+1}) +
    \lambda \mathcal{L}_{\text{reg}}(\Z),
\end{equation}
Here, $\mathcal{L}_{\text{pred}}$ denotes the latent-state prediction error
(e.g., mean squared error), 
and $\lambda$ is the regularization weight.

\section{Experiments}\label{sec:exp}
\subsection{Experimental Setup}
\label{subsec:exp-setup}

\paragraph{Environments.}
We evaluate \name{} on four continuous-control benchmarks:
Two-Room~\cite{zhou2025dino-wm}, a 2D navigation task with low intrinsic
dimensionality; Reacher~\cite{Reacher}, a two-link planar reaching task;
PushT~\cite{chi2025diffusion}, a 2D block-pushing manipulation task; and
OGB-Cube~\cite{park2025ogbench}, a visually rich 3D manipulation environment.
All tasks have continuous action spaces, and all models are trained
directly from raw RGB observations.

\paragraph{Baselines.}
We compare \name{} against three representative baselines.
LeWM~\cite{maes_lelidec2026lewm} is our primary baseline and the
direct foundation of our method. 
It imposes a global Gaussian constraint
in the ambient embedding space.
PLDM~\cite{sobal2025stresstesting} is an end-to-end pixel-based
world model trained with multiple heuristic objectives.
DINO-WM~\cite{zhou2025dino-wm} uses a frozen pretrained DINOv2~\cite{oquab2024dinov}
visual encoder to mitigate representation collapse. For a fair
comparison with pixel-only end-to-end world models, we report
DINO-WM without proprioceptive inputs as the main reference and
include the proprioceptive variant only as an additional
upper-bound reference when available.

\paragraph{Implementation Details.}
\name{} uses the same encoder, latent predictor, optimizer, training
schedule, and loss-weight tuning protocol as LeWM, except that
full-space Gaussian regularization is replaced by Multi-Subspace Gaussian regularization
(~\cref{subsec:multisig}).
The embedding dimension is set to $D=192$ for all experiments.
We use $K=32$ for all environments except PushT where we instead use $K=16$ for better performance.
The values are selected through ablation studies on
a held-out validation 
set, as described in~\cref{subsec:ablation}.
The projection matrices $\{\proj_k\}$ are orthogonally initialized and
kept frozen throughout training, as described in
~\cref{subsec:osd}.

\subsection{Planning Performance}
\label{subsec:planning}

\subsubsection{Success Rates}
We first report the planning success rates. 
Overall, 
\name{} consistently outperforms LeWM~\cite{maes_lelidec2026lewm} across all four environments, 
demonstrating that relocating the Gaussian regularization into subspaces 
yields more powerful latent representations.

For a fair comparison, 
LeWM and \name{} are evaluated across six random seeds (reporting mean $\pm$ std). 
Since official checkpoints for PLDM~\cite{sobal2025stresstesting} and DINO-WM~\cite{zhou2025dino-wm} are unavailable, we cite their success rates directly from LeWM under the exact same protocol.

\begin{table*}[!htb]
\centering
\caption{
Planning success rate (\%, higher is better) across four
continuous-control environments.
\name{} consistently improves upon LeWM across all tasks,
with the largest gain on
Two-Room~\cite{zhou2025dino-wm},
where the full-space Gaussian prior of LeWM is most mismatched
to the low intrinsic dimensionality of the task.
}
\label{tab:planning}

\begin{tabular}{lcccc}
\toprule
\textbf{Method}
& \textbf{Two-Room~\cite{zhou2025dino-wm}}
& \textbf{Reacher~\cite{Reacher}}
& \textbf{PushT~\cite{chi2025diffusion}}
& \textbf{OGB-Cube~\cite{park2025ogbench}} \\
\midrule

PLDM~\cite{sobal2025stresstesting}
& 97.00
& 78.00
& 78.00
& 65.00 \\

DINO-WM (w/o proprio.)~\cite{zhou2025dino-wm}
& 100.00
& 79.00
& 74.00
& 86.00 \\

DINO-WM (w/ proprio.)~\cite{zhou2025dino-wm}
& 100.00
& --
& 92.00
& -- \\

LeWM~\cite{maes_lelidec2026lewm}
& 84.33{\tiny$\pm$4.23}
& 82.67{\tiny$\pm$4.42}
& 84.67{\tiny$\pm$6.53}
& 67.33{\tiny$\pm$5.01} \\

\name{} (Ours)
& 95.00{\tiny$\pm$2.76}
& 84.00{\tiny$\pm$4.00}
& 89.00{\tiny$\pm$5.33}
& 76.33{\tiny$\pm$5.99} \\

\bottomrule
\end{tabular}
\end{table*}

As shown in~\cref{tab:planning},
the improvement is most prominent on Two-Room~\cite{zhou2025dino-wm}. LeWM is known to struggle here because enforcing isotropic Gaussianity over the entire high-dimensional embedding introduces an overly restrictive bias for this low-intrinsic-dimension task. By contrast, \name{} regularizes lower-dimensional orthogonal views. This relaxes the global constraint and better preserves task-relevant structure while maintaining the anti-collapse benefits. 

On Reacher~\cite{Reacher} and PushT~\cite{chi2025diffusion}, \name{} achieves strong performance, consistently outperforming PLDM and surpassing DINO-WM without relying on a pretrained visual backbone. On OGB-Cube~\cite{park2025ogbench}, \name{} substantially improves upon LeWM. Although DINO-WM still leads in visually complex 3D manipulation tasks (where pretrained features excel), \name{} provides a highly effective, simple end-to-end alternative without relying on external pretraining.

\subsubsection{Success Rates w.r.t. Effective Rank}
To verify whether our subspace regularization yields more compact latent representations as motivated, we analyze the effective rank~\cite{roy2007effective} of the learned embeddings.
Effective rank measures how many principal directions are effectively used by the latent embeddings.
%

For each environment, we uniformly sample $N=2000$ observations from
the evaluation set and use the same sampled observations for LeWM and
\name{}. Each observation is fed through the trained encoder with
weights frozen, and the projected \texttt{[CLS]} embedding is used as
the latent representation. 
Let $Z \in \mathbb{R}^{N \times D}$ denote the centered latent
matrix, and let $\{\lambda_i\}_{i=1}^{D}$ be the eigenvalues
of its empirical covariance matrix. We normalize the covariance
spectrum as $p_i = \lambda_i / \sum_j \lambda_j$ and define the
effective rank as
\begin{equation}
r_{\mathrm{eff}} =
\exp\left(-\sum_i p_i \log p_i\right).
\end{equation}

\begin{figure}[t]
\centering
\begin{tikzpicture}
\begin{groupplot}[
    group style={
        group size=1 by 2,
        vertical sep=0.3cm,
    },
    width=0.8\linewidth,
    height=4.0cm,
    xmin=0.7, xmax=4.3,
    ylabel style={
        font=\small
    },
    yticklabel style={
        text width=1.2em,
        align=right
    },
    xtick={1,2,3,4},
    xticklabels={Two-Room, OGB-Cube, PushT, Reacher},
    xticklabel style={
        font=\small
    },
    tick align=outside,
    ytick pos=left,
    xtick pos=bottom,
    axis line style={line width=0.6pt},
    tick style={line width=0.6pt},
    legend style={
        draw=none,
        fill=none,
        font=\small,
        at={(0.98,0.98)},
        anchor=north east,
        legend columns=1,
    },
    every axis plot/.append style={
        line width=1.4pt,
        mark size=1.4pt,
    },
]

\nextgroupplot[
    ylabel={Effective Rank},
    ymin=0, ymax=90,
    ytick={0,40,80},
    yticklabel style={
        font=\small
    },
    xticklabel=\empty,
]

\addplot[
    color=ourscolor,
    mark=*,
]
coordinates {
    (1,7.25)
    (2,5.70)
    (3,6.13)
    (4,37.84)
};
\addlegendentry{\name}

\addplot[
    color=lewmcolor,
    dashed,
    mark=square*,
    mark options={solid},
]
coordinates {
    (1,79.50)
    (2,40.82)
    (3,7.11)
    (4,42.26)
};
\addlegendentry{LeWM}

\nextgroupplot[
    ylabel={Success Rate (\%)},
    ymin=60, ymax=100,
    ytick={60,80,100},
    yticklabel style={
        font=\small
    },
    xticklabel style={
        rotate=0,
        anchor=north,
    },
]
    
\addplot[
    color=ourscolor,
    mark=*,
]
coordinates {
    (1,95.00)
    (2,76.33)
    (3,89.00)
    (4,84.00)
};

\addplot[
    color=lewmcolor,
    dashed,
    mark=square*,
    mark options={solid},
]
coordinates {
    (1,84.33)
    (2,67.33)
    (3,84.67)
    (4,82.67)
};

\end{groupplot}
\end{tikzpicture}
\vspace{-1em}
\caption{
Effective rank and planning success rate across 
four environments. 
Top: effective rank of latent representations. Bottom: planning success
rate.
Larger effective-rank reductions from LeWM to \name{} broadly
coincide with larger planning gains, suggesting that Multi-Subspace
Gaussian regularization improves performance by reducing spurious
high-rank variation in the latent space.
}
\label{fig:r_eff-SR}
\end{figure}
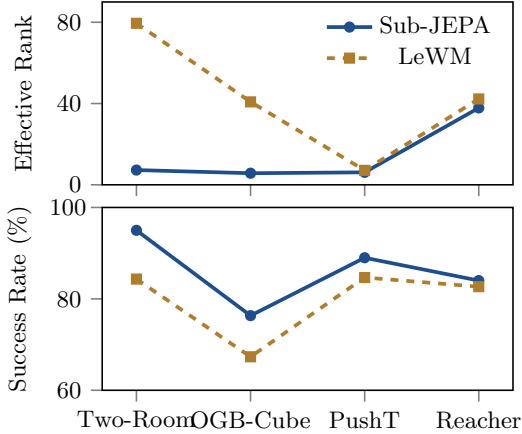

As shown in~\cref{fig:r_eff-SR}, the effective-rank reduction from LeWM
to \name{} is strongly and directly correlated with planning improvement
across all four environments. 
Two-Room~\cite{zhou2025dino-wm} and
OGB-Cube~\cite{park2025ogbench} exhibit the largest rank compressions
and simultaneously the largest success-rate gains, while
PushT~\cite{chi2025diffusion} and Reacher~\cite{Reacher} show smaller
rank gaps and more moderate improvements.

This direct correspondence strongly supports our hypothesis: when
task-relevant features lie on a low-dimensional manifold, the full-space
Gaussian prior in LeWM forces the latent space into unnecessarily high
rank, imposing a bias that hinders task-aligned representation learning.
Subspace-wise regularization relaxes this bias and allows the latent
geometry to contract toward the task's intrinsic dimensionality---the
greater this contraction, the larger the planning gain.


\subsection{Ablation Studies}\label{subsec:ablation}
We conduct ablation studies to validate the design choices of \name{} 
and to understand the bias-variance tradeoff in subspace design.

\subsubsection{Number of subspaces $K$}
We first study the effect of the number of subspaces $K$. 
Specifically, we set $D=192$ and vary
$K \in \{1, 2, 4, 8, 16, 32, 64\}$, 
which corresponds to subspace dimensions $\ds = \{192, 96, 48, 24, 12, 6,3\}$.
When $K=1$ and $\ds=D$, the regularizer reduces to a full-space
Gaussian regularization with an orthogonal transformation, 
making it the closest counterpart to LeWM~\cite{maes_lelidec2026lewm} within our implementation, 
whereas larger $K$ values decompose the
same latent space into progressively smaller subspaces.

\begin{table}[!htb]
\centering
\caption{
    Planning success rate (\%) for different numbers of subspaces
    $K$ under fixed total latent dimensionality $D=192$, where
    $\ds=192/K$.
    Gray cells indicate configurations that outperform the
    LeWM baseline under the same evaluation protocol.
    Subspace-wise regularization improves performance across a broad
    range of configurations.
}
\label{tab:ablation-k}

\ifdefined\ieeemode
\else
\setlength{\tabcolsep}{3pt}
\fi

\begin{tabular}{ccccc}
\toprule
\textbf{$K$}
& \textbf{Two-Room}
& \textbf{Reacher}
& \textbf{PushT}
& \textbf{OGB-Cube} \\
\midrule
\color{gray}LeWM & \color{gray}{84.33{\tiny$\pm$4.23}} & \color{gray}{82.67{\tiny$\pm$4.42}} & \color{gray}{84.67{\tiny$\pm$6.53}} & \color{gray}{67.33{\tiny$\pm$5.01}} \\
1
& \cellcolor{rowgray}87.00{\tiny$\pm$4.43}
& 82.00{\tiny$\pm$6.32}
& 84.67{\tiny$\pm$5.47}
& \cellcolor{rowgray}68.00{\tiny$\pm$6.20} \\

2
& \cellcolor{rowgray}91.33{\tiny$\pm$4.42}
& 80.67{\tiny$\pm$6.39}
& \cellcolor{rowgray}88.00{\tiny$\pm$7.04}
& \cellcolor{rowgray}68.00{\tiny$\pm$5.40} \\

4
& \cellcolor{rowgray}92.00{\tiny$\pm$2.83}
& \cellcolor{rowgray}83.67{\tiny$\pm$3.88}
& \cellcolor{rowgray}88.67{\tiny$\pm$4.76}
& \cellcolor{rowgray}70.00{\tiny$\pm$6.45} \\

8
& \cellcolor{rowgray}91.00{\tiny$\pm$3.29}
& \cellcolor{rowgray}82.68{\tiny$\pm$4.38}
& \cellcolor{rowgray}85.67{\tiny$\pm$2.94}
& \cellcolor{rowgray}70.33{\tiny$\pm$5.89} \\

16
& \cellcolor{rowgray}93.67{\tiny$\pm$4.27}
& 81.00{\tiny$\pm$2.10}
& \cellcolor{rowgray}89.00{\tiny$\pm$5.33}
& \cellcolor{rowgray}69.00{\tiny$\pm$8.69} \\

32
& \cellcolor{rowgray}95.00{\tiny$\pm$2.76}
& \cellcolor{rowgray}84.00{\tiny$\pm$4.00}
& 28.00{\tiny$\pm$5.22}
& \cellcolor{rowgray}76.33{\tiny$\pm$5.99} \\

64
& \cellcolor{rowgray}92.00{\tiny$\pm$2.34}
& 74.33{\tiny$\pm$7.53}
& 9.00{\tiny$\pm$5.33}
& \cellcolor{rowgray}67.67{\tiny$\pm$6.12} \\

\bottomrule
\end{tabular}
\end{table}

Results are reported in ~\cref{tab:ablation-k}. The effect of
$K$ is task-dependent. On Two-Room~\cite{zhou2025dino-wm},
performance improves as $K$ increases from small to moderate values,
peaking around $K=32$, after which it slightly degrades. This is
consistent with the low-dimensional structure of the task:
relaxing the full-space Gaussian constraint allows the model to
capture the underlying navigation dynamics more faithfully. A similar trend is observed on OGB-Cube~\cite{park2025ogbench},
where larger $K$ consistently improves performance and achieves
the best result at $K=32$. Reacher~\cite{Reacher} also benefits
slightly from moderate increases in $K$, suggesting that
subspace-wise regularization remains effective even in visually
complex or continuous-control settings.

A different trend is observed on PushT~\cite{chi2025diffusion}.
In PushT, performance collapses at
$K=32$, where each subspace contains only six dimensions.
This suggests that excessively aggressive partitioning can make
individual subspaces too low-dimensional to provide stable and
informative normality estimates\ifdefined\ieeemode\else~\cite{alemi2017deep}\fi, 
particularly for manipulation
tasks requiring tightly coupled object--agent interactions.

Overall, $K$ controls a trade-off between flexibility and
statistical reliability. Increasing $K$ relaxes the global
Gaussian prior and enables more expressive latent structure,
while excessively small subspaces may weaken the effectiveness
of the regularization signal.

\subsubsection{Joint effect of $K$ and $\ds$}
In this ablation,
we relax the constraint $\ds = D/K$ 
and independently vary both $K$ and $\ds$ 
to study their joint effect on planning performance. 

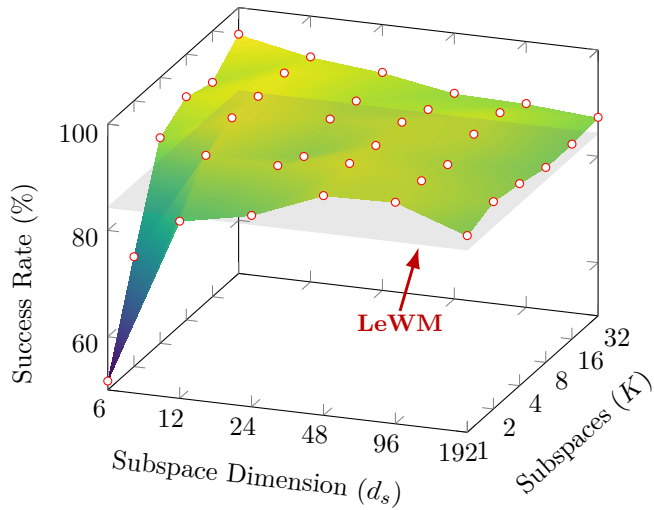
\begin{figure}[t]
\centering
\begin{tikzpicture}
\begin{axis}[
    view={-70}{25},
    width=0.92\linewidth,
    height=7.2cm,
    xlabel={Subspaces ($K$)},
    ylabel={Subspace Dimension ($\ds$)},
    zlabel={Success Rate (\%)},
    xtick={1,2,3,4,5,6},
    xticklabels={1,2,4,8,16,32},
    ytick={1,2,3,4,5,6},
    yticklabels={192,96,48,24,12,6},
    zmin=50,
    zmax=100,
    xlabel style={sloped like x axis},
    ylabel style={sloped like y axis},
    colormap/viridis,
]

\addplot3[
    surf,
    shader=interp,
    mesh/cols=6
] coordinates {
(1,1,87.00) (2,1,89.00) (3,1,88.00) (4,1,86.67) (5,1,86.67) (6,1,87.33)

(1,2,91.67) (2,2,91.33) (3,2,90.00) (4,2,91.33) (5,2,91.00) (6,2,88.33)

(1,3,91.33) (2,3,93.00) (3,3,92.00) (4,3,92.00) (5,3,90.00) (6,3,88.67)

(1,4,86.00) (2,4,91.00) (3,4,88.33) (4,4,91.00) (5,4,90) (6,4,91)

(1,5,83.33) (2,5,91.33) (3,5,94) (4,5,93.67) (5,5,93.67) (6,5,92.33)

(1,6,51.67) (2,6,70.67) (3,6,88.67) (4,6,92.00) (5,6,90.33) (6,6,95)
};
\addplot3[
    surf,
    shader=flat,
    draw=none,
    opacity=0.18,
    samples=2,
    domain=1:6,
    y domain=1:6,
    color=gray,
]
{84.33};
\addplot3[
    only marks,
    mark=*,
    mark size=1.5pt,
    mark options={draw=red,fill=white}
] coordinates {
(1,1,87.00) (2,1,89.00) (3,1,88.00) (4,1,86.67) (5,1,86.67) (6,1,87.33)

(1,2,91.67) (2,2,91.33) (3,2,90.00) (4,2,91.33) (5,2,91.00) (6,2,88.33)

(1,3,91.33) (2,3,93.00) (3,3,92.00) (4,3,92.00) (5,3,90.00) (6,3,88.67)

(1,4,86.00) (2,4,91.00) (3,4,88.33) (4,4,91.00) (5,4,90) (6,4,91)

(1,5,83.33) (2,5,91.33) (3,5,94) (4,5,93.67) (5,5,93.67) (6,5,92.33)

(1,6,51.67) (2,6,70.67) (3,6,88.67) (4,6,92.00) (5,6,90.33) (6,6,95)
};

\draw[
    -{Latex[length=2.8mm,width=2mm]},
    red!75!black,
    line width=1.1pt
]
(rel axis cs:0.86,0.50) -- (rel axis cs:1.36,0.63);

\node[
    text=red!75!black,
    font=\small\bfseries,
    anchor=north,
    inner sep=1pt
]
at (rel axis cs:0.86,0.50) {LeWM};

\end{axis}
\end{tikzpicture}
\vspace{-2em}
\caption{
    Success rate on Two-Room~\cite{zhou2025dino-wm} is shown as a function of $K$ and $\ds$,
    with the baseline success rate of LeWM~\cite{maes_lelidec2026lewm} shown as a flat reference plane.
    \name{} outperforms LeWM across a broad 
    mid-range of configurations.
}
\label{fig:subspace-surface}
\end{figure}

As shown in~\cref{fig:subspace-surface}, performance is stable across
a broad mid-range of $(K, d_s)$ configurations, but degrades sharply
when subspaces become too narrow (small $d_s$), where the normality
signal grows unreliable. This reveals a bias-variance tradeoff in
subspace design: larger $K$ relaxes the global prior but requires
sufficient $d_s$ to maintain regularization stability.

\subsubsection{Justification of frozen orthogonal projection}

We use orthogonally initialized frozen projection matrices
and ablate this choice against two alternatives:
(1) randomly initialized frozen projection, where the projection
matrices are sampled from a standard Gaussian and kept fixed;
and (2) trainable projection with soft orthogonality regularization,
where the matrices are initialized orthogonally but updated
during training subject to an orthogonality penalty.

\begin{figure*}
    \centering
    \begin{overpic}[width=0.95\linewidth]{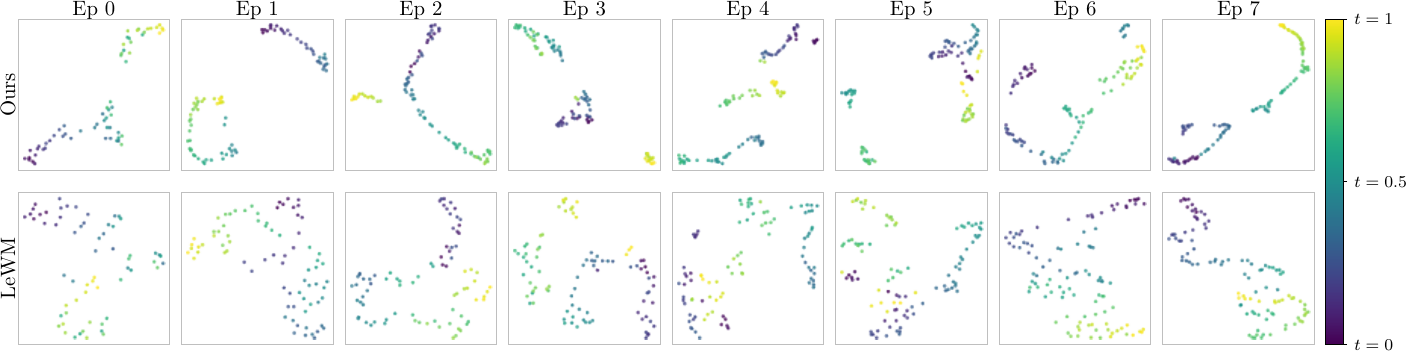}
    \end{overpic}
    \vspace{-1em}
    \caption{
        Visualization of latent embeddings extracted from consecutive observations in representative Two-Room~\cite{zhou2025dino-wm} episodes.
        Each column corresponds to one episode, with Sub-JEPA shown in the top row and LeWM shown in the bottom row. 
        Each point denotes one time step, 
        and colors indicate the normalized timestamp. 
        Sub-JEPA produces more temporally coherent latent trajectories, while LeWM exhibits less organized temporal evolution in several episodes. 
    }
\label{fig:latent_traj_umap}
\end{figure*}

As shown in~\cref{tab:ablation-proj},
orthogonal frozen projection consistently achieves the best
performance across all environments. This confirms that the
projection design is not merely an implementation detail, 
but a key
component of \name. 
By preserving geometric
isometry, the orthogonal frozen projection ensures that each
subspace receives a balanced and non-redundant view of the latent
representation, allowing the subspace-level normality constraints to
act uniformly~\cite{halko2011finding}.

\begin{table}[!t]
\centering
\caption{
    Planning success rate (\%) across environments for different
    projection strategies.
    Orthogonal frozen projection gives the most stable and effective
    subspace projection.
}
\label{tab:ablation-proj}
\setlength{\tabcolsep}{1.5pt}
\ifdefined\ieeemode
\else
\resizebox{\linewidth}{!}{%
\fi
\begin{tabular}{lcccc}
\toprule
\textbf{Projection}
    & \textbf{Two-Room} & \textbf{Reacher}
    & \textbf{PushT} & \textbf{OGB-Cube} \\
\midrule
Ortho init~\frozen
    & \textbf{95.00{\tiny$\pm$2.76}}
    & \textbf{84.00{\tiny$\pm$4.00}}
    & \textbf{89.00{\tiny$\pm$5.33}}
    & \textbf{76.33{\tiny$\pm$5.99}} \\
Rand init~\frozen
    & 53.00{\tiny$\pm$8.44}
    & 68.00{\tiny$\pm$5.29}
    & 13.33{\tiny$\pm$5.61}
    & 61.00{\tiny$\pm$5.55} \\
Ortho reg~\trainable
    & 61.67{\tiny$\pm$5.16}
    & 82.67{\tiny$\pm$4.68}
    & 57.00{\tiny$\pm$9.27}
    & 70.33{\tiny$\pm$7.09} \\
\bottomrule
\end{tabular}
\ifdefined\ieeemode
\else
}
\fi
\end{table}

Random frozen projection performs substantially worse, especially on
PushT~\cite{chi2025diffusion}. 
Without orthogonality, different subspaces may
contain redundant or unevenly scaled information, weakening the
intended projection of the full latent space.
Trainable projection with soft orthogonality regularization also
underperforms the frozen orthogonal variant. We attribute this to
co-adaptation between the encoder and the projection matrices: as
training progresses, the learned projections can align with
directions that reduce the effective strength of the regularizer,
thereby weakening the anti-collapse effect of Multi-Subspace Gaussian regularization.

\subsection{Physical State Probing}
\label{subsec:physical-probing}

Planning success evaluates whether a learned world model supports
goal-directed control, but it does not directly characterize the
physical information contained in the latent representation.
We therefore evaluate the physical decodability of the learned
representations on PushT~\cite{chi2025diffusion}.
Following the probing protocol of LeWM~\cite{maes_lelidec2026lewm},
we freeze the encoder and train lightweight probes from latent
embeddings to ground-truth physical variables.
We consider three task-relevant quantities: agent location, block
location, and block angle.
For each quantity, we train both a linear probe and a small MLP probe.
The linear probe measures whether the variable is directly accessible
from the representation, while the MLP probe tests whether the
information is retained in a possibly nonlinear form.
We report mean squared error (MSE) and Pearson correlation coefficient
$r$, averaged over six random seeds.

\cref{tab:physical-probing} reports the physical latent probing results on PushT.
Overall, \name{} matches or outperforms LeWM across most properties and probe types,
suggesting that Multi-Subspace Gaussian regularization preserves physically meaningful
latent structure while improving nonlinear recoverability.

An interesting exception arises for block angle prediction under the linear probe, where
\name{} slightly underperforms LeWM. We hypothesize that rotation is an inherently
non-linear quantity---subspace projection may fragment the angular structure across
independent subspaces, making it less linearly decodable. Consistent with this view,
the gap closes under the MLP probe, suggesting that a nonlinear decoder can compensate
for this fragmentation. This pattern highlights a nuanced tradeoff: subspace
regularization improves the linear decodability of translational quantities while
potentially complicating that of rotational ones.

\begin{table}[t]
\centering
\small
\setlength{\tabcolsep}{3.5pt}
\caption{Physical latent probing results on PushT.}
\label{tab:physical-probing}
\begin{tabular}{llcccc}
\toprule
\textbf{Property} & \textbf{Model}
& \multicolumn{2}{c}{\textbf{Linear}}
& \multicolumn{2}{c}{\textbf{MLP}} \\
\cmidrule(lr){3-4}\cmidrule(lr){5-6}
& & \textbf{MSE $\downarrow$} & \textbf{$r \uparrow$}
& \textbf{MSE $\downarrow$} & \textbf{$r \uparrow$} \\
\midrule

\multirow{3}{*}{Agent Loc.}
& PLDM
    & 0.090
    & 0.955
    & 0.014
    & 0.993 \\
& LeWM
    & 0.052
    & 0.974
    & 0.004
    & 0.998 \\
& \cellcolor{rowgray}\textbf{Ours}
    & \cellcolor{rowgray}\textbf{0.048}
    & \cellcolor{rowgray}\textbf{0.976}
    & \cellcolor{rowgray}\textbf{0.002}
    & \cellcolor{rowgray}\textbf{0.999} \\
\midrule

\multirow{3}{*}{Block Loc.}
& PLDM
    & 0.122
    & 0.938
    & 0.011
    & 0.994 \\
& LeWM
    & 0.029
    & 0.986
    & \textbf{0.001}
    & 0.999 \\
& \cellcolor{rowgray}\textbf{Ours}
    & \cellcolor{rowgray}\textbf{0.024}
    & \cellcolor{rowgray}\textbf{0.988}
    & \cellcolor{rowgray}\textbf{0.001}
    & \cellcolor{rowgray}\textbf{1.000} \\
\midrule

\multirow{3}{*}{Block Angle}
& PLDM
    & 0.446
    & 0.745
    & 0.056
    & 0.972 \\
& LeWM
    & \textbf{0.187}
    & \textbf{0.902}
    & 0.022
    & 0.989 \\
& \cellcolor{rowgray}\textbf{Ours}
    & \cellcolor{rowgray}0.218
    & \cellcolor{rowgray}0.884
    & \cellcolor{rowgray}\textbf{0.021}
    & \cellcolor{rowgray}\textbf{0.990} \\
\bottomrule
\end{tabular}
\end{table}

\subsection{Latent Trajectories}
\label{sec:latent-trajectory-visualization}

To verify that \name{} learns representations better matched to the
task geometry, we visualize latent trajectories on Two-Room~\cite{zhou2025dino-wm},
where the low intrinsic dimensionality makes the mismatch with the
full-space Gaussian prior most pronounced.
For each episode, consecutive observations are encoded and the
\texttt{[CLS]} embeddings are projected to 2D via UMAP~\cite{healy2024uniform},
colored by normalized temporal index.

As shown in~\cref{fig:latent_traj_umap}, \name{} consistently
produces temporally coherent latent trajectories across episodes,
with temporally adjacent observations forming well-organized paths.
LeWM, by contrast, exhibits less regular temporal structure in
several episodes, suggesting that the full-space Gaussian prior
distorts latent geometry when task dynamics are intrinsically
low-dimensional.

\subsection{Temporal Latent Path Straightening}\label{subsec:straightening}

Following LeWM~\cite{maes_lelidec2026lewm}, we examine latent trajectory
geometry via temporal path straightening, which measures how linearly
dynamics evolve in latent space~\cite{henaff2019straightening,interno2026aigeneratedvideodetectionperceptual}---a
property particularly relevant for latent world models where planning
quality depends on rollout regularity rather than pixel
reconstruction~\cite{ha2018world,Hansen2022tdmpc,lecun2022path}.

\begin{figure*}[!htb]
\hfill
\begin{overpic}[width=0.95\linewidth]{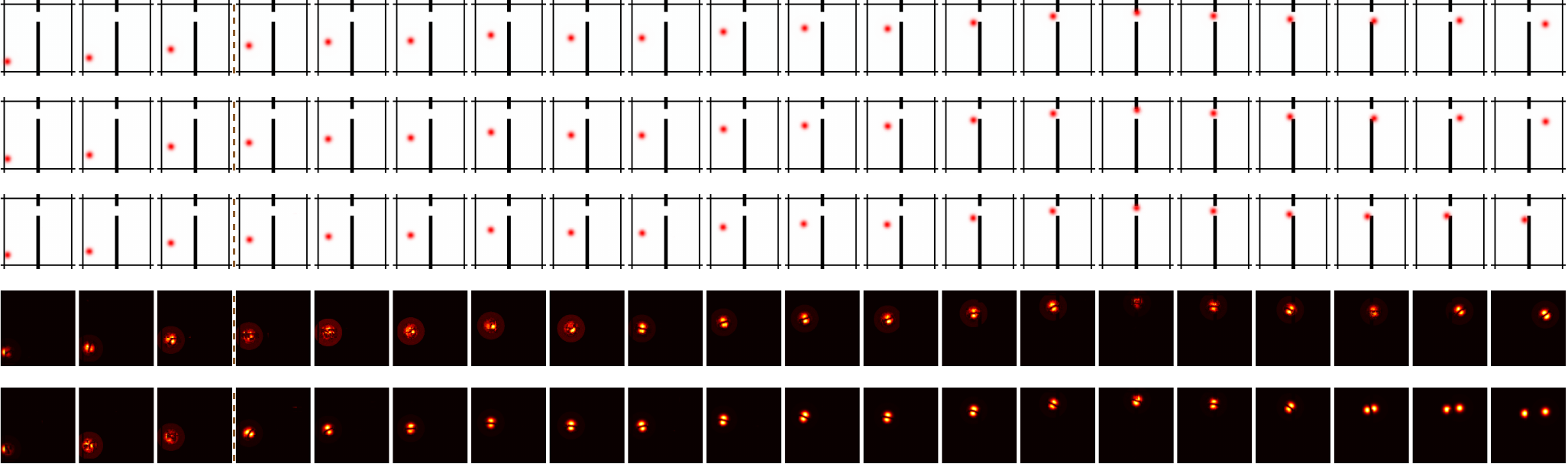}
    \put(3.5, -1.5){\footnotesize $t=5$}
    \put(15.8,-1.5){\footnotesize 15}
    \put(26.4,-1.5){\footnotesize 25}
    \put(37,-1.5){\footnotesize 35}
    \put(46.2,-1.5){\footnotesize 45}
    \put(56.6,-1.5){\footnotesize 55}
    \put(66.8,-1.5){\footnotesize 65}
    \put(76.4,-1.5){\footnotesize 75}
    \put(86.4,-1.5){\footnotesize 85}
    \put(96.3, -1.5){\footnotesize 95}
    \put(14.85, 0){\rule{0.8pt}{5.5cm}}
    \put(1, 30){Context input}
    \put(46, 30){Open-Loop predictions}
    \put(-4.3, 12.5){\rotatebox{90}{\textbf{\color{blue}{Predictions}}}}
    \put(-3, 25.5){\rotatebox{90}{GT}}
    \put(-2.2, 19){\rotatebox{90}{Ours}}
    \put(-2, 12.5){\rotatebox{90}{\footnotesize{LeWM}}}
    \put(-4.3, 3){\rotatebox{90}{\textbf{\color{red}{Error}}}}
    \put(-2.2, 6.5){\rotatebox{90}{Ours}}
    \put(-2, 0.2){\rotatebox{90}{\footnotesize{LeWM}}}
\end{overpic}
\vspace{0.5em}
\caption{
    Open-loop rollout comparison between \name{} and LeWM on Two-Room~\cite{zhou2025dino-wm}.
    Models are conditioned on three context frames ($t=\{0,5,10\}$) and then
    recursively predict future observations using only actions. Rows 2--3 show
    predicted trajectories for \name{} and LeWM, respectively. 
    Rows 4--5 show
    absolute pixel errors relative to the ground truth. \name{} exhibits more
    accurate long-horizon predictions and reduced cumulative drift.
}\label{fig:openloop-rollout}
\end{figure*}
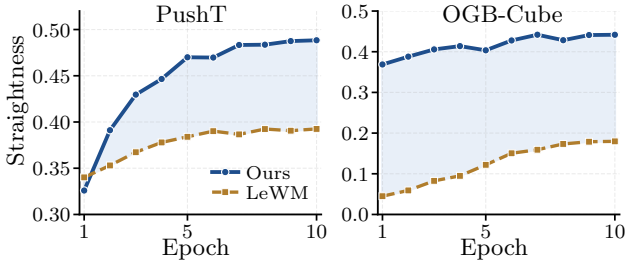
\begin{figure}[!htb]
\centering
\begin{overpic}[width=0.9\linewidth]{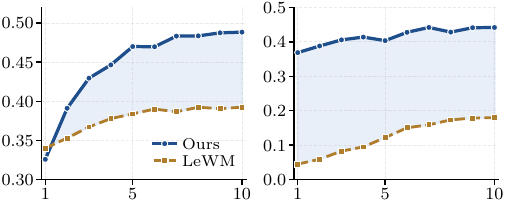}
\put(21, 37){\small PushT}
\put(69, 37){\small OGB-Cube}
\put(22, -2){\small Epoch}
\put(72, -2){\small Epoch}
\put(-4, 13){\rotatebox{90}{\small Straightness}}
\end{overpic}
\caption{
    Temporal latent path straightening over training, measured by mean cosine similarity between consecutive latent velocity vectors (higher = straighter trajectories). \name{} achieves consistently higher temporal straightness than LeWM on both environments, suggesting that Multi-Subspace Gaussian regularization implicitly encourages smoother latent dynamics.
}
\label{fig:temporal-straightening}
\end{figure}
Given latent embeddings $z_{1:T} \in \mathbb{R}^{B \times T \times D}$, we define the temporal velocity as $v_t = z_{t+1} - z_t$, and measure straightness via the mean cosine similarity between consecutive velocities~\cite{maes_lelidec2026lewm}:
\begin{equation}
S_{\mathrm{straight}}
=
\frac{1}{B(T-2)}
\sum_{i=1}^{B}
\sum_{t=1}^{T-2}
\frac{
\left\langle v^{(i)}_t, v^{(i)}_{t+1} \right\rangle
}{
\left\|v^{(i)}_t\right\|
\left\|v^{(i)}_{t+1}\right\|
} .
\end{equation}
Higher values indicate smoother and more linearly evolving latent trajectories.

As shown in~\cref{fig:temporal-straightening}, \name{} consistently
produces straighter latent trajectories than LeWM on both
PushT~\cite{chi2025diffusion} and OGB-Cube~\cite{park2025ogbench},
emerging naturally without explicit optimization. This suggests that
subspace-wise regularization reduces geometric distortion relative to
full-space SIGReg, offering a geometric explanation for the planning
gains in~\cref{tab:planning}.

\subsection{Open-loop Rollout Visualization}
\label{openloop}

To evaluate long-horizon stability, we compare open-loop rollouts of
\name{} and LeWM~\cite{maes_lelidec2026lewm} on
Two-Room~\cite{zhou2025dino-wm}. Each model is conditioned on three
context frames at $t=\{0,5,10\}$ and autoregressively predicts future
states from actions alone, decoded into RGB over 20 steps up to $t=95$.
As shown in~\cref{fig:openloop-rollout}, both models produce accurate
short-horizon predictions, but diverge at longer horizons. LeWM
exhibits increasing spatial drift and structural distortion, while
\name{} maintains scene geometry more consistently, yielding lower
reconstruction error at distant steps.

This suggests that subspace-wise regularization improves recursive
latent stability: by replacing a single global Gaussian constraint
with structured subspace constraints, \name{} better supports
coherent long-term state propagation.

\section*{Conclusion}\label{sec:conclusion}
We presented \name, 
a simple yet effective extension of 
LeWorldModel that relocates Gaussian regularization 
from the ambient embedding space 
into random orthogonal subspaces.
Our key insight is that the isotropic Gaussian prior enforced by 
LeWM operates in the full high-dimensional latent space, which 
introduces an excessive bias when the underlying task dynamics 
reside on a low-dimensional manifold.
Extensive experiments demonstrate that \name{} consistently
outperforms LeWM across four tasks, that physical state information
is better preserved and more nonlinearly decodable in the learned
representations, and that the latent trajectories are substantially
more temporally coherent.

\bibliographystyle{unsrt}
\bibliography{eg}
\end{document}